\let\oldvec\vec
\documentclass[runningheads,a4paper]{llncs}
\let\vec\oldvec
\usepackage{times}  

\usepackage{amssymb}
\usepackage{amsmath}
\makeatletter
\g@addto@macro\normalsize{%
  \setlength\abovedisplayskip{-10pt}
  \setlength\belowdisplayskip{-5pt}
  \setlength\abovedisplayshortskip{-10pt}
  \setlength\belowdisplayshortskip{-5pt}
}
\makeatother
\usepackage{graphicx}
\usepackage{url}
\usepackage{xcolor}
\usepackage{verbatim}
\usepackage[parfill]{parskip}
\usepackage{algpseudocode,algorithm}
\usepackage[bookmarks=true,pdfpagelabels=false,plainpages=true]{hyperref}
\usepackage{subcaption}
\hypersetup{
    unicode=false,          
    pdftoolbar=true,        
    pdfmenubar=true,        
    pdffitwindow=false,     
    pdfstartview={FitH},    
    pdfnewwindow=true,      
    colorlinks=true,       
    filecolor=magenta,      
    linkcolor={red!50!black},
    citecolor={blue!50!black},
    urlcolor={blue!80!black},
    bookmarksopen=true,
    bookmarksnumbered=true
}
\usepackage[all]{hypcap}

\makeatletter
\newcommand\thefontsize[1]{{#1 The current font size is: \f@size pt\par}}
\makeatother

\algnewcommand\algorithmicinput{\textbf{Input:}}
\algnewcommand\Input{\item[\algorithmicinput]}
\algnewcommand\algorithmicoutput{\textbf{Output:}}
\algnewcommand\Output{\item[\algorithmicoutput]}

\renewcommand\textfloatsep{4pt}

\renewcommand\intextsep{4pt}

\begin{document}
\mainmatter
\title{Improving Tweet Representations using Temporal and User Context}

\author{Ganesh J\inst{1} \and  Manish Gupta\inst{1,2} \and Vasudeva Varma\inst{1}}
\institute{IIIT, Hyderabad, India\\
\email{ganesh.j@research.iiit.ac.in, vv@iiit.ac.in}
\and
Microsoft, India\\
\email{gmanish@microsoft.com}
}
\maketitle

\begin{abstract}
In this work we propose a novel representation learning model which computes semantic representations for tweets accurately. Our model systematically exploits the chronologically adjacent tweets (`context') from users' Twitter timelines for this task. Further, we make our model user-aware so that it can do well in modeling the target tweet by exploiting the rich knowledge about the user such as the way the user writes the post and also summarizing the topics on which the user writes. We empirically demonstrate that the proposed models outperform the state-of-the-art models in predicting the user profile attributes like spouse, education and job by 19.66\%, 2.27\% and 2.22\% respectively.
\end{abstract}

\section{Introduction} 
\label{sec:intro}

\begin{figure*}[ht]
\centering
\includegraphics[height=2.25cm,width=0.75\columnwidth]{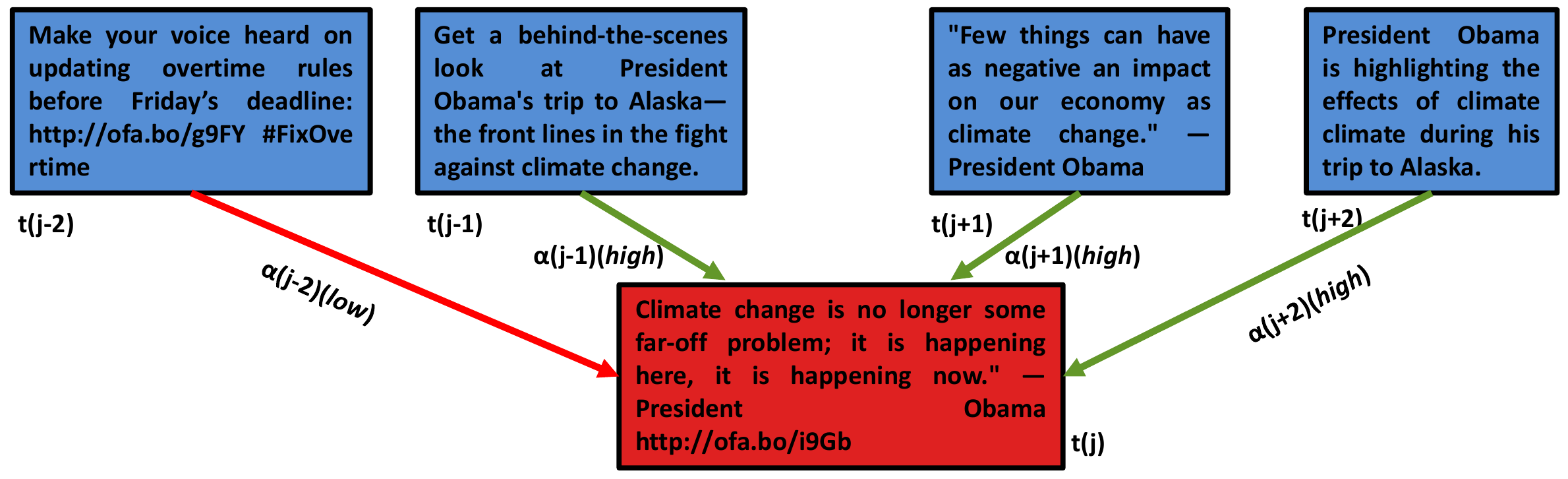}
\caption{$t(j-1)$, $t(j-2)$, $t(j+1)$ and $t(j+2)$ form the temporal context of $t(j)$. $\alpha$'s denote the attention parameters of the proposed model.}
\label{fig:at_fig}
\end{figure*}

The short and noisy nature of tweets poses challenges in computing accurate latent tweet representations. We observe that Paragraph2Vec~\cite{le14_icml} which is good in computing document representation overfits when evaluated for tweets, mainly due to the short length of tweets. To overcome this problem we utilize additional context from Twitter itself. Specifically, we hypothesize that a principled usage of chronologically adjacent tweets from users' Twitter timelines can help in significantly improving the quality of the representation. The main challenge lies in assigning appropriate attention weights to context tweets such that semantically relevant tweets receive high weights compared to less relevant ones. Consider Fig~\ref{fig:at_fig}\footnote{The tweets are borrowed from Barack Obama's Twitter timeline posted in Sep 2015.}, where we want to learn the representation for the tweet $t(j)$. One can see that the target tweet $t(j)$ has less semantic interactions with the context tweet $t(j-2)$. To capture this, we propose an attention based model that assigns a variable weight to each context tweet that captures the semantic correspondence between the target tweet and the context tweet. We further augment the attention model to be user-aware so that it can do well in modeling the target tweet by exploiting the rich knowledge about the user such as the way the user writes the post, and also summarizing the topics on which the user writes. Our work is closest to~\cite{djuric15_www} where documents are modeled based on their word context as well as document stream context. We differ from their work in two ways: (1) they na\"{\i}vely assume that all the documents in a stream have equal amount of semantic interactions and, (2) they ignore the knowledge of user (or document author).

We summarize our main contributions below.
In summary, our contributions are as follows. (1) Our work is the first to model the semantics of the tweet using the temporal context. 
(2) We introduce a novel attention based model that learns the weights for context tweets by back-propagating semantic loss. 
(3) We propose a novel way to learn user vector summarizing the content the user writes, which in turn helps in enriching the quality of the tweet embeddings. 
(4) We conduct quantitative analysis to showcase the application potential of the tweet representations learned from the model and also provide some interesting findings.


\section{Related Work} 
\label{sec:related}
Le et al.~\cite{le14_icml} adapt Word2Vec to learn document representations which are good in predicting the words present in the document. As seen in Section~\ref{sec:experiments}, for short documents like tweets, the model tends to learn poor document representations as the vector relies too much on the document content, resulting in overfitting. Djuric et al.~\cite{djuric15_www} learn document representations using word context (same as~\cite{le14_icml}) along with document stream context in a hierarchical fashion. This work inspired us to learn tweet representations using user specific Twitter streams.

\section{Problem Formulation} 
\label{sec:problem}
In this section we first introduce the notions of temporal context and attention, and then provide a formal problem statement. 

\noindent\underline{\textbf{Temporal context}}: Temporal context of a tweet $t(j)$ is the set of $C_T$ tweets posted before and after $t(j)$ by the same user. The value $C_T$ is a user specified parameter that defines the size of the temporal context to be considered to model a given tweet. For example, in Fig~\ref{fig:at_fig} we fix $C_T$ as 2, the context tweets of $t(j)$ are $t(j-1)$, $t(j-2)$, $t(j+1)$ and $t(j+2)$. 

\noindent\underline{\textbf{Attention}}: An attention value is associated with a context tweet that defines the degree of semantic similarity between the context tweet and the target tweet. The more the latent semantic interactions between the tweets, the more is the attention. We denote the attention of context tweet $t(j-1)$ as $\alpha(j-1)$. For instance, in Fig~\ref{fig:at_fig}, the attention value of context tweet $t(j-2)$ should be lower than that of context tweet $t(j-1)$ with respect to target tweet $t(j)$. In Fig~\ref{fig:at_fig}, clearly $t(j-2)$ is not talking about the topic `Climate Change' and so it makes sense to have a lower attention value. 

\noindent\underline{\textbf{Problem Statement}}: Let the training tweets be given in the order in which they are posted. In particular, we assume that we are given a user set $U$ of $N_u$ tweet sequences, with each sequence $u(k) \in U$, containing $N_t$ tweets, $u(k)=\{t(1),..,t(j),..,t(N_t)\}$ posted by user $u(k)$. Moreover, each tweet $t(j)$ is a sequence of $N_w$ words, $t(j)=\{w(j,1),..,w(j,i),..,w(j,N_w)\}$. The problem is to learn semantic low-dimensional representations for all the tweets in the sequences in set $U$.

\section{Proposed Models} 
\label{sec:approach}

\begin{figure*}
    \centering
    \begin{subfigure}[t]{0.5\textwidth}
        \centering
        \includegraphics[height=0.8in, width=2in]{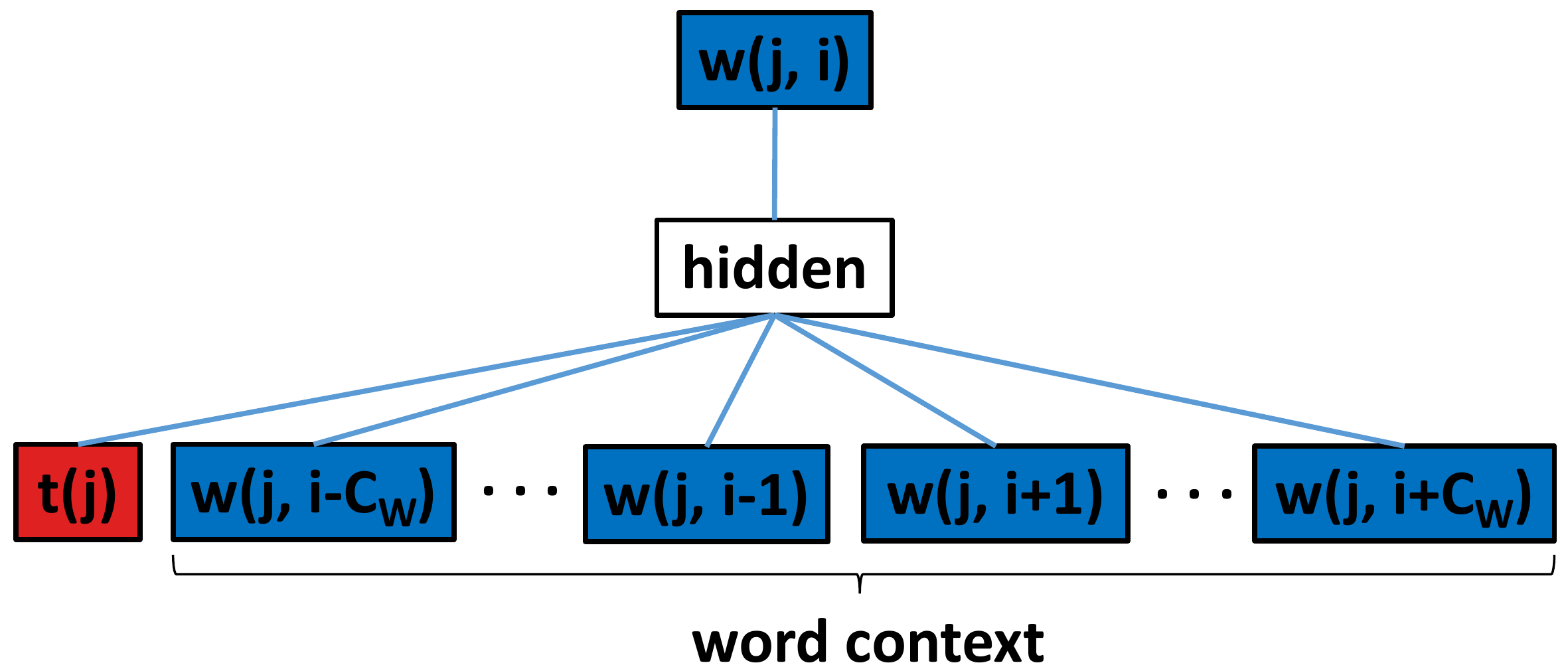}
        \caption{Word Context Model}
    \end{subfigure}%
    ~ 
    \begin{subfigure}[t]{0.5\textwidth}
        \centering
        \includegraphics[height=0.8in, width=2in]{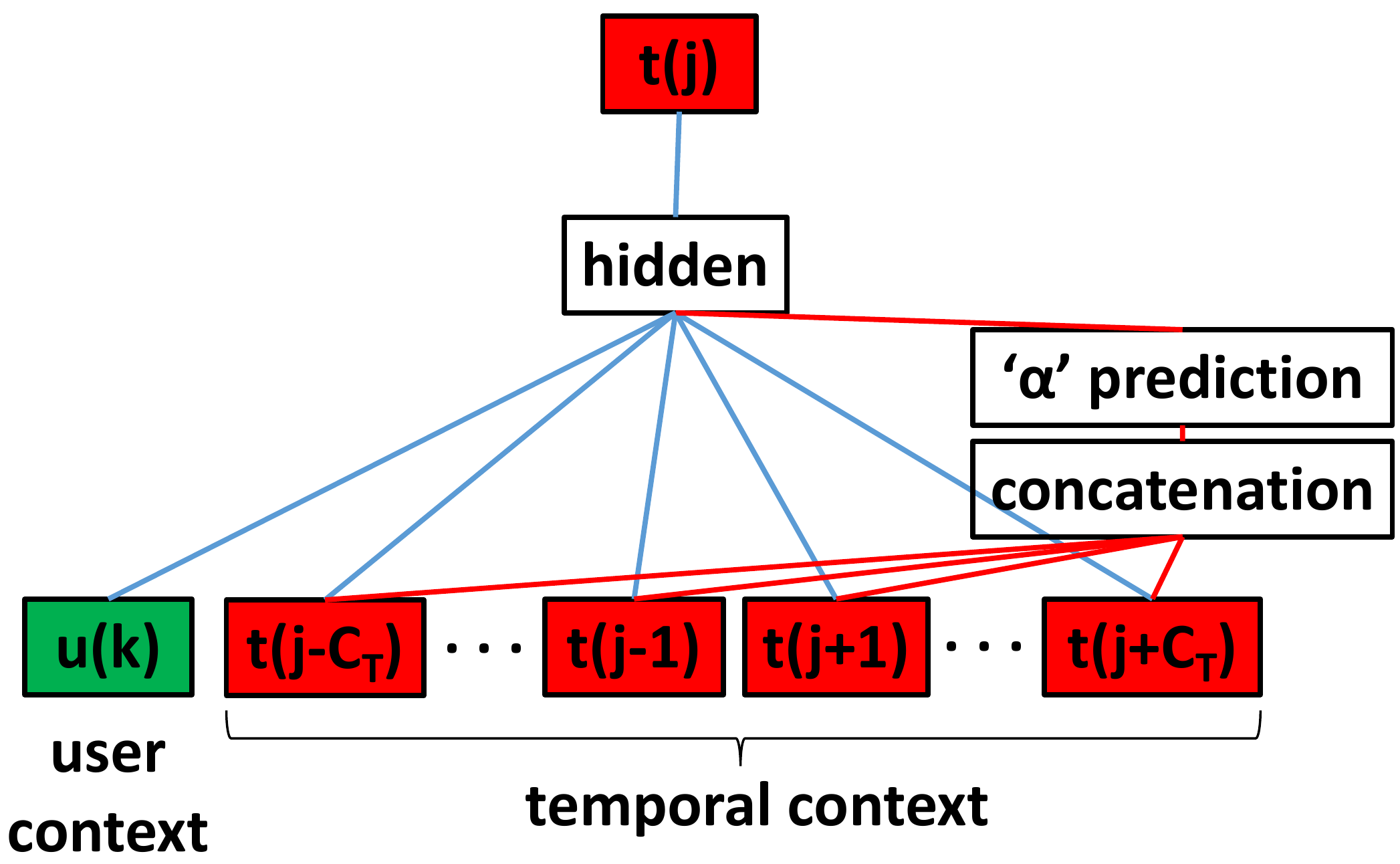}
        \caption{User + Tweet Context Model}
    \end{subfigure}
    \caption{Architecture diagram of our model.}
    \label{fig:t2v_fig}
\end{figure*}


Our model (Fig~\ref{fig:t2v_fig}) learns tweet representations in a hierarchical fashion: learning from the words present in the tweet using word context model (Fig~\ref{fig:t2v_fig} (a)) along with the temporal tweets present in the user stream using tweet context model (Fig~\ref{fig:t2v_fig} (b)). Both the models will be discussed in detail below. Let $\textbf{w(j,i)}$, $\textbf{t(j)}$ and $\textbf{u(k)}$ denote the embedding for a word $i$ from tweet $j$, tweet $j$ and user $u(k)$ respectively, all of which have the size `n'. We will discuss details about both of these models in this section.

\subsection{Word Context Model}


The goal of the word context model is to learn tweet representations which are good at predicting the words present in the tweet. The model has three layers. The first layer contains the word embeddings, $\bf{w(j, i-C_W)} , \cdots, \textbf{w(j, i-1)}$, $\textbf{w(j, i+1)}, \cdots, \bf{w(j, i+C_W)}$ near the $i^{th}$ target word in tweet $j$, which denote the word context for the word $i$ (i.e., $w(j, i)$) along with the tweet embedding $\textbf{t(j)}$. Secondly, there is a hidden layer with size equal to the number of words in the vocabulary ($|V|$). The final layer is a softmax layer which gives a well-defined probability distribution over words in the vocabulary. The input to the word context model is all pairs of word context of word $i$ and tweet $t(j)$ in the corpus. The objective is to maximize the likelihood of the word $w(j, i)$ occurring given its context, i.e., $\mathbb{P}(w(j, i) | w(j, i-C_W), \cdots , w(j, i-1), w(j, i+1), \cdots, w(j, i+C_W),t(j))$. Equation~\ref{eq:eq2} represents the forward propagation step in our 1-hidden layer feed forward model, where $W_{WC}$ and $T_{WC}$ denote the additional parameters of the model.


\scriptsize
\begin{eqnarray}
{\hat{y}}_{|V|\times1}(j)
=
softmax(  
W_{WC} \times
\sum_{l \in \{ i-C_W, i+C_W) \} \setminus i }  \textbf{w(j, l)}
+ 
T_{WC} \times \textbf{t(j)}
)
\label{eq:eq2}
\end{eqnarray}
\normalsize


\subsection{User + Tweet Context Model}

The goal of this model is to enrich the tweet representation learned from the word context, by modeling the current tweet conditioned on its temporal context and the proposed user context. The user context makes our model user-aware by exploiting the user characteristics such as the way the user writes the post and also summarizing the topics on which the user writes. These user vectors are learned automatically from the set of tweets posted by the user through this model. As a na\"{i}ve solution, we can directly adopt Djuric et al.~\cite{djuric15_www}'s approach and apply on the Twitter stream. As discussed in Section~\ref{sec:problem}, this assumption is too strong for social media streams. Can we assign attention levels to the context tweets with respect to the tweet being modeled? To learn the optimal values of attention ($\alpha(j)$), we introduce the attention parameters as shown in Equation~\ref{eq:eq5}. The intuition is that semantic loss will be less if the weights of each of the temporal context tweets are learned accurately. The values of $\alpha(j)$'s can be computed as shown in Equation~\ref{eq:eq6}. The objective of this model is to maximize the likelihood of the tweet $j$ posted by user $k$ given its temporal context ($\textbf{t($\bf{j-C_T}$)}, \cdots, \textbf{t(j-1)}, \textbf{t(j+1)} , \cdots, \textbf{t(j+$\bf{C_T}$)}$) and user context ($\textbf{u(k)}$), which is given by $\mathbb{P}(t(j) | t(j-C_T), \cdots, t(j-1), t(j+1), \cdots, t(j+C_T), u(k))$. Since the tweet space can be exponentially large, we use hierarchical softmax~\cite{morin05_aistats} instead of normal softmax to bring down the time complexity from $O(|T|)$ (or $O(|V|)$ for the previous model) to $O(log|T|)$ (or $O(log|V|$)).

\scriptsize
\begin{eqnarray}
{\hat{y}}_{|T| \times 1}(j)
=
softmax(
T_{TC} \times
\sum_{l \in \{ j-C_T, j+C_T \} \setminus j}  \alpha(l) \times \textbf{t(l)}
)
\label{eq:eq5}
\end{eqnarray}
\normalsize




\scriptsize
\begin{eqnarray}
\label{eq:eq6}
(\alpha(j-C_T) \cdots \alpha(j-1) \alpha(j+1) \cdots \alpha(j+C_T) ) = softmax (A [ \textbf{t($\bf{j-C_T}$)}; \cdots; \textbf{t(j-1)}; \textbf{t(j+1)}; \cdots;\textbf{t(j+$\bf{C_T}$)}; ] )
\end{eqnarray}

\normalsize



\noindent where the parenthesis inside the softmax function represents concatenation of all context representations (($2 \times C_T \times n)\times 1$ in size). $A$ is the additional weight matrix (of size $({2 \times C_T})\times({2 \times C_T \times n})$) added as parameters to the model. In practice, we observe that multiple passes (`epochs') on the training set are required to fine tune these attention values. The overall objective function intertwining both the models in a hierarchical fashion to be maximized can be summarized as shown in Equation~\ref{eq:eq7}. We use the cross-entropy as the cost function between the predicted distribution $\hat{y}(j)$ and target distributions $t(j)$ and $w(j,i)$, for modeling using the temporal and word context respectively. We train the model using back-propagation~\cite{rumelhart86_nature} and Adam~\cite{kingma_arxiv} optimizer.




\scriptsize
\begin{eqnarray}
\nonumber \mathcal{L}(\theta)=\sum\limits_{u(k) \in U} \biggl[ \sum\limits_{t(j) \in u(k)} \sum\limits_{w(j,i) \in t(j)} \log\mathbb{P}(w(j,i) | w(j,i-C_W) ,\cdots, w(j,i-1), w(j,i+1), \cdots \\
\nonumber , w(j,i+C_W), t(j)) + \log \mathbb{P}(t(j) | w(j,1), \cdots, w(j,N_w)) + \log \mathbb{P}(t(j) | t(j-C_T), \cdots, t(j-1), t(j+1), \\
\cdots, t(j+C_T), u(k)) \biggr] + \log \mathbb{P}(u(k) | t(1), \cdots, t(N_T))
\label{eq:eq7}
\end{eqnarray}
\normalsize


\section{Experimental Evaluation} 
\label{sec:experiments}

In this section we discuss details of our dataset, experiment, and then present quantitative analysis of the proposed models. 


\begin{table*}[ht]
\scriptsize
\centering
\begin{tabular}{|c|c|c|c|}
\hline
\textbf{Algorithm} & \textbf{Spouse} & \textbf{Education} & \textbf{Job} \\ \hline
Paragraph2Vec~\cite{le14_icml} & 0.3435 & 0.9259 & 0.5465 \\ \hline
Simple Distance model (SD) & 0.3704 & 0.9068 & 0.5872 \\ \hline
HDV~\cite{djuric15_www} & 0.4526 & 0.8901 & 0.521 \\ \hline
Ours (User = 0) & \textbf{0.5416} & 0.9098 & 0.5935 \\ \hline
Ours (User = 1) & 0.4082 & \textbf{0.9274} & \textbf{0.6067} \\ \hline
\end{tabular}
\caption{User profile attribute classification - F1 Score}
\label{tab:classaccuracy}
\end{table*}

\subsection{Dataset Description}
We use the publicly available dataset described in Li et al.~\cite{li14_acl} for all the experiments. It contains tweets pertaining to three profile attributes (spouse, education and job) of a user. Specifically, it has a set of tweets from users' Twitter timelines, that talk about the attribute (`positive' tweets) and those that do not (`negative' tweets). We randomly sample 1600 users from the dataset and use 70-10-20 ratio to construct train, validation and test splits. Tweet embeddings are randomly initialized while the word embeddings are initialized with the pre-trained word vectors from Pennington et al.~\cite{pennington14_emnlp}. 

\subsection{Experimental Protocol}

We consider the binary task of predicting whether a given entity mention corresponds to particular users' profile attribute or not. We build our model to get the tweet vector and the entity vector by computing an average of all the tweet vectors for the entity. We tune the penalty parameter of a linear Support Vector Machine (SVM) on the validation set. Note that we use a linear classifier so as to minimize the effect of variance of non-linear methods on the classification performance and subsequently help in interpreting the results. We compare our model with three baselines: (1) Paragraph2Vec~\cite{le14_icml}, (2) Simple Distance model (SD): A model that assigns attention weight to the context tweet which is inversely proportional to the distance of the tweet from the target tweet, (3) HDV~\cite{djuric15_www}, (4) Ours (User = 0): Our model when the user context is excluded from the temporal context, (5) Ours (User = 1): Our model when the user context is included in the temporal context.  We empirically set $n$ and $C_W$ to 200 and 10 respectively for all the models. In case of SD, HDV and our models, we try values in \{1, 2, 4, 6, 8, 10, 12, 14, 16\} to fix the temporal context size parameter (i.e., $C_T$) which is crucial in improving the semantics of the tweet. 

\begin{figure*}[t!]
    \centering
    \begin{subfigure}[t]{0.3\textwidth}
        \centering
        \includegraphics[height=0.6in, width=1.5in]{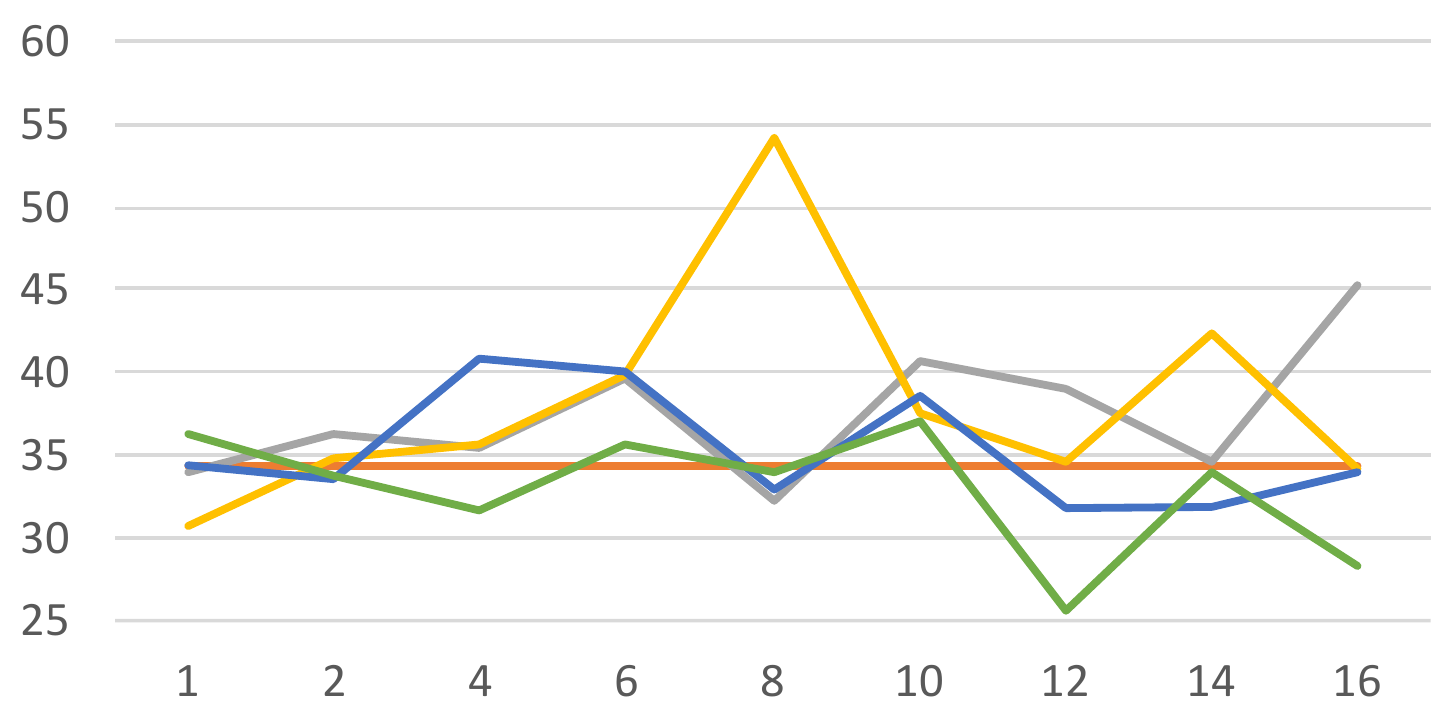}
        \caption{Spouse}
    \end{subfigure}%
    ~ 
    \begin{subfigure}[t]{0.29\textwidth}
        \centering
        \includegraphics[height=0.6in, width=1.5in]{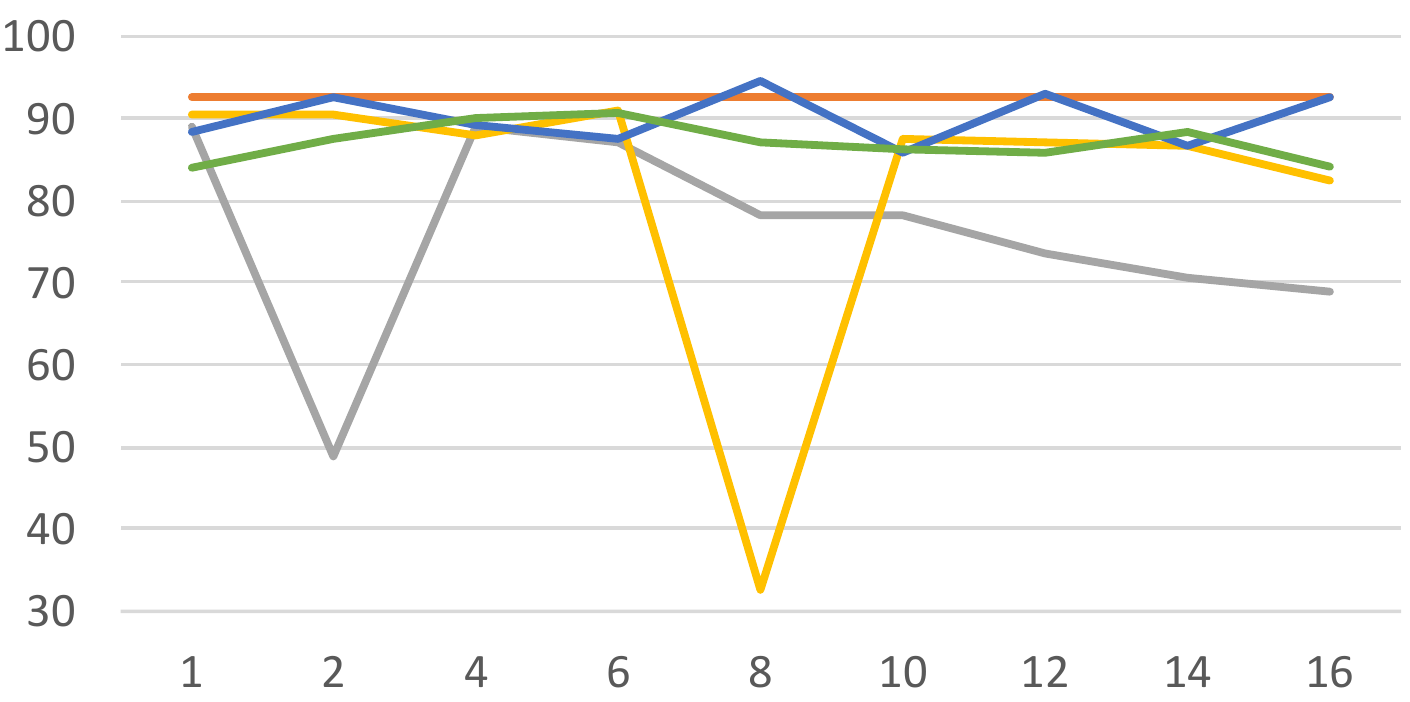}
        \caption{Education}
    \end{subfigure}
    ~
    \begin{subfigure}[t]{0.29\textwidth}
        \centering
        \includegraphics[height=0.6in, width=1.5in]{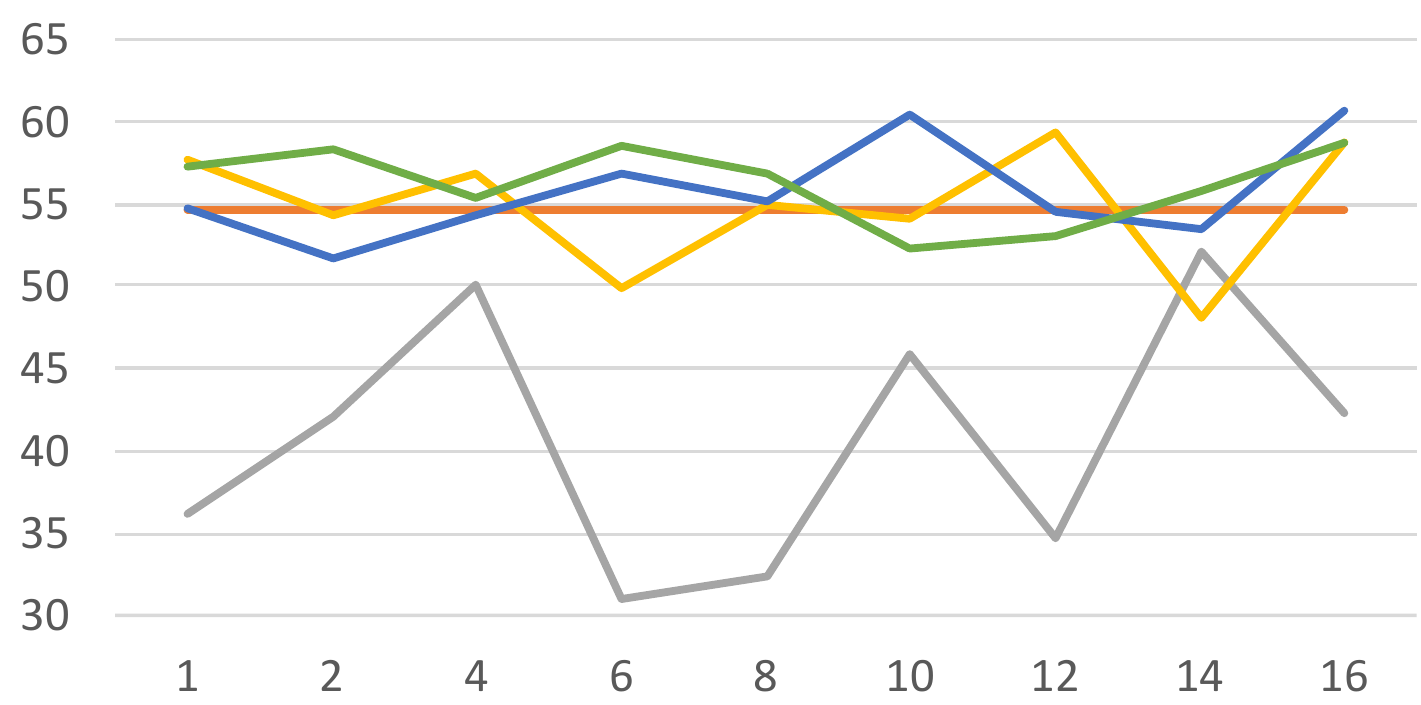}
        \caption{Job}
    \end{subfigure}
    \begin{subfigure}[t]{1\textwidth}
        \centering
        \includegraphics[height=0.1in]{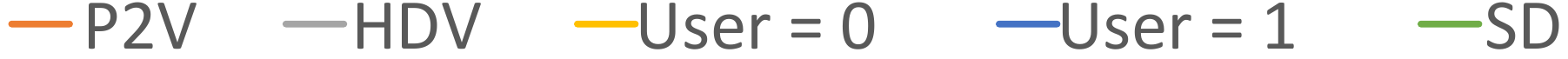}
    \end{subfigure}
    \caption{Model performance w.r.t. temporal context size $C_T$.}
    \label{fig:perf_ct}
\end{figure*}

\begin{figure*}[t!]
    \centering
    \begin{subfigure}[t]{0.33\textwidth}
        \centering
        \includegraphics[height=0.6in, width=1.5in]{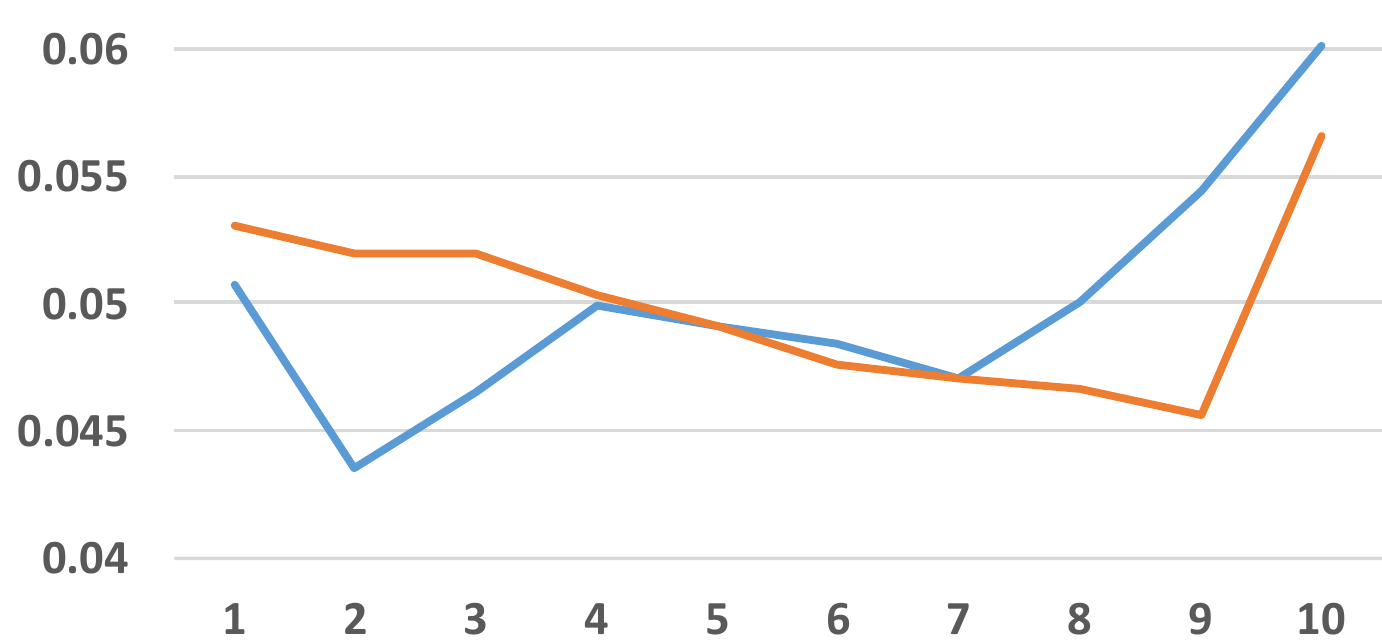}
        \caption{Spouse}
    \end{subfigure}%
    ~ 
    \begin{subfigure}[t]{0.29\textwidth}
        \centering
        \includegraphics[height=0.6in, width=1.5in]{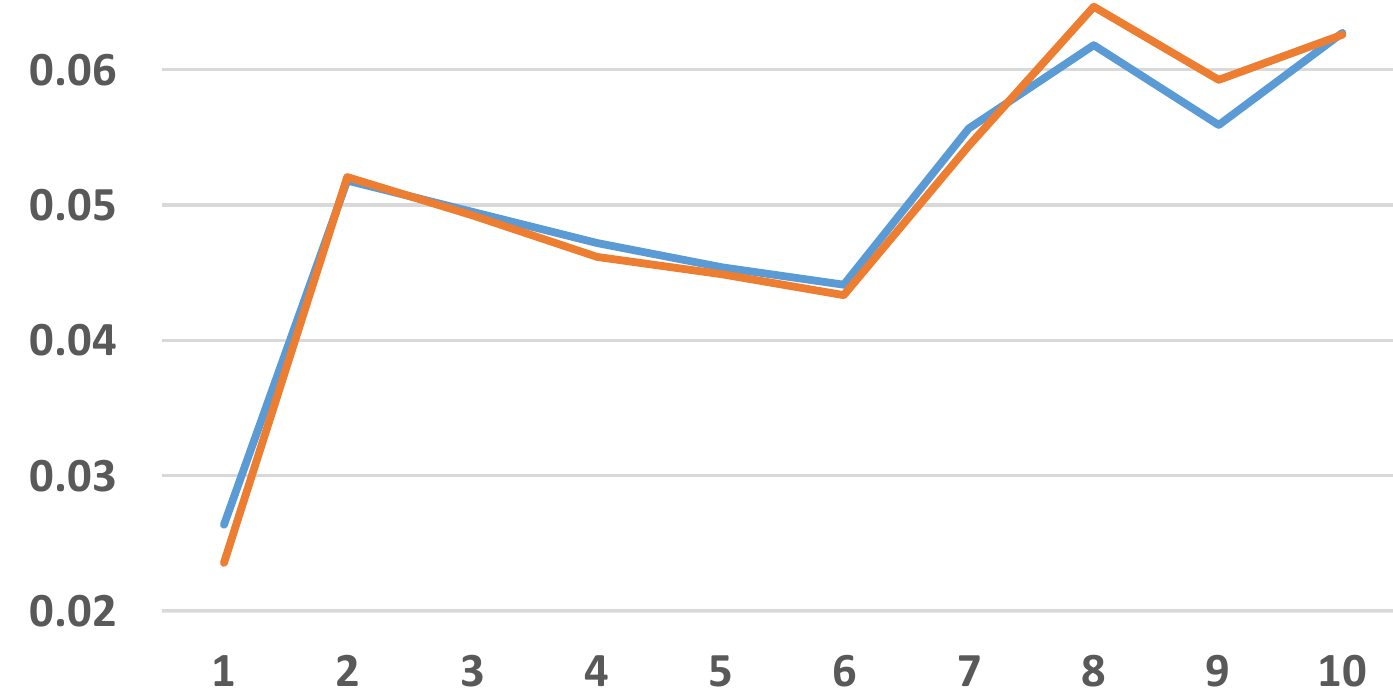}
        \caption{Education}
    \end{subfigure}
    ~
    \begin{subfigure}[t]{0.29\textwidth}
        \centering
        \includegraphics[height=0.6in, width=1.5in]{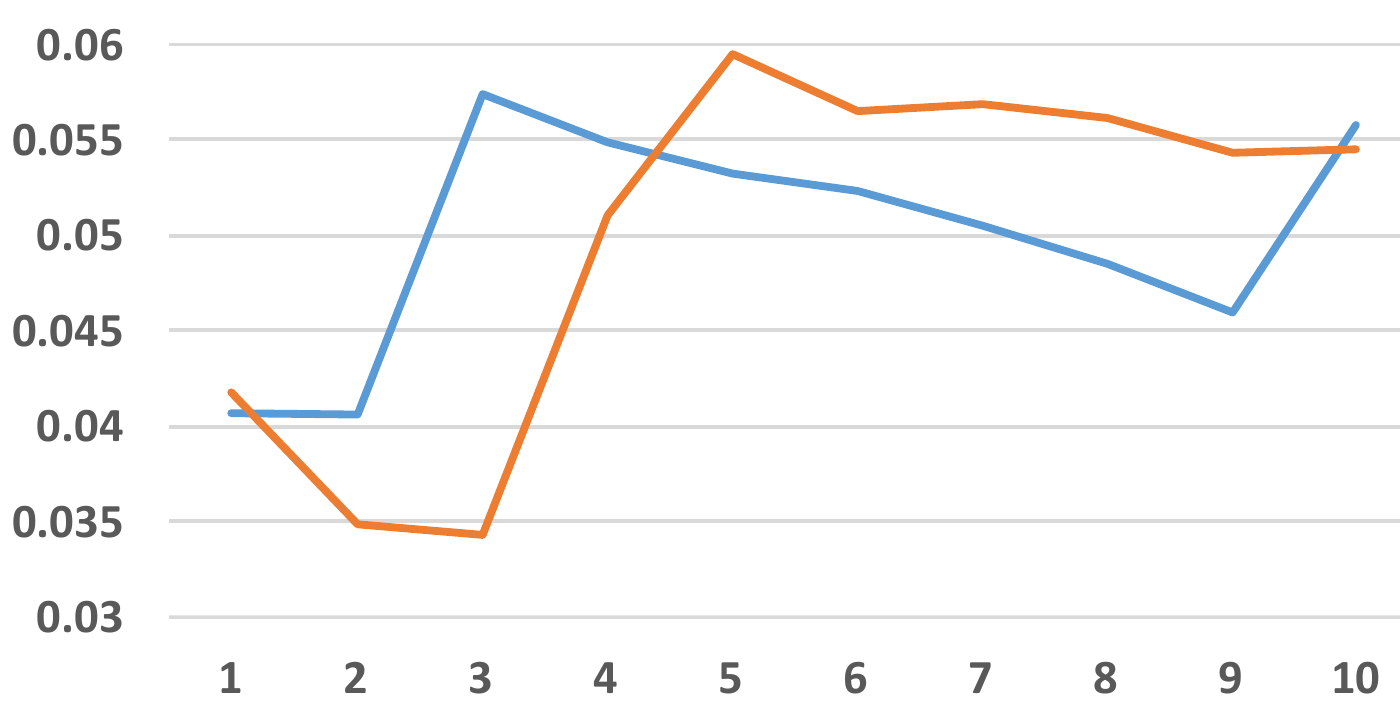}
        \caption{Job}
    \end{subfigure}
    \begin{subfigure}[t]{1\textwidth}
        \centering
        \includegraphics[height=0.1in]{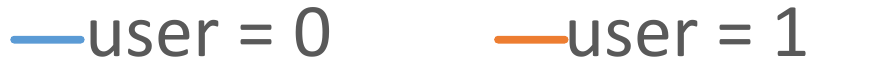}
    \end{subfigure}
    \caption{Mean attention w.r.t. distance from the center $C_T$.}
    \label{fig:att_dist}
\end{figure*}

\subsection{Comparative analysis}

From Table~\ref{tab:classaccuracy}, we see that Paragraph2Vec overfits the validation set, resulting in poor accuracy during testing. HDV's assumption of giving equal attention value to the temporal context also results in lower accuracy compared with our models. SD model outperforms HDV in two tasks, which substantiates our claim against HDV's na\"{\i}ve assumption for social media. Our model with user vector outperforming the baselines for Education and Job attribute classification, shows the need to consider the user characteristics while modelling his/her tweets. The poor results for Spouse task suggest that this dataset has too many topic shifts and that the user vector turned out to be less accurate. Fig~\ref{fig:perf_ct} displays the F1 results for different values of $C_T$, which is a vital parameter controlling the influence of temporal context. We observe that in some cases HDV outperforms the SD model, mainly due to the inability of the SD model to utilize the context information from farther tweets which are relevant with respect to the target tweet. Our models are 19.66\%, 2.27\% and 2.22\% better compared to the baselines for the spouse, education and job attributes respectively.

\subsection{Impact of Variable Attention}

\begin{figure*}[t!]
    \centering
    \begin{subfigure}[t]{0.3\textwidth}
        \centering
        \includegraphics[height=0.6in, width=2in]{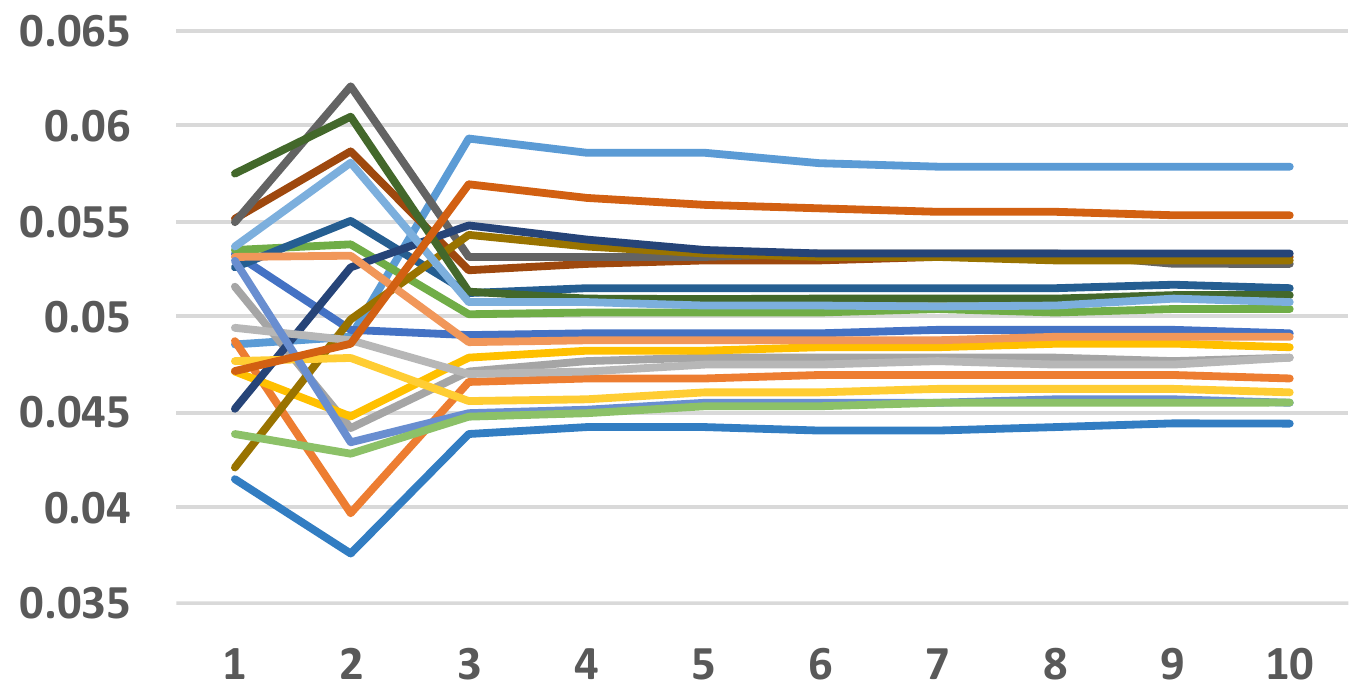}
        \caption{Spouse}
    \end{subfigure}
    ~ 
    \begin{subfigure}[t]{0.3\textwidth}
        \centering
        \includegraphics[height=0.6in, width=2in]{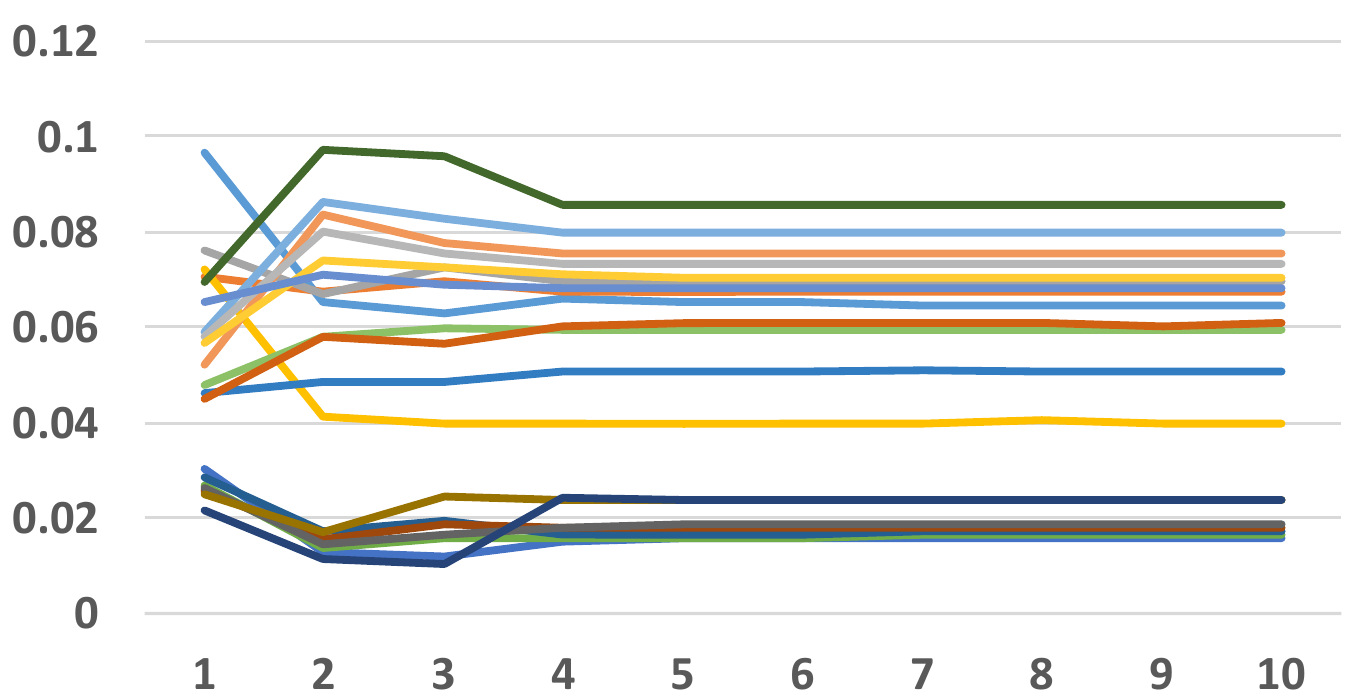}
        \caption{Education}
    \end{subfigure}
    ~
    \begin{subfigure}[t]{0.3\textwidth}
        \centering
        \includegraphics[height=0.6in, width=2in]{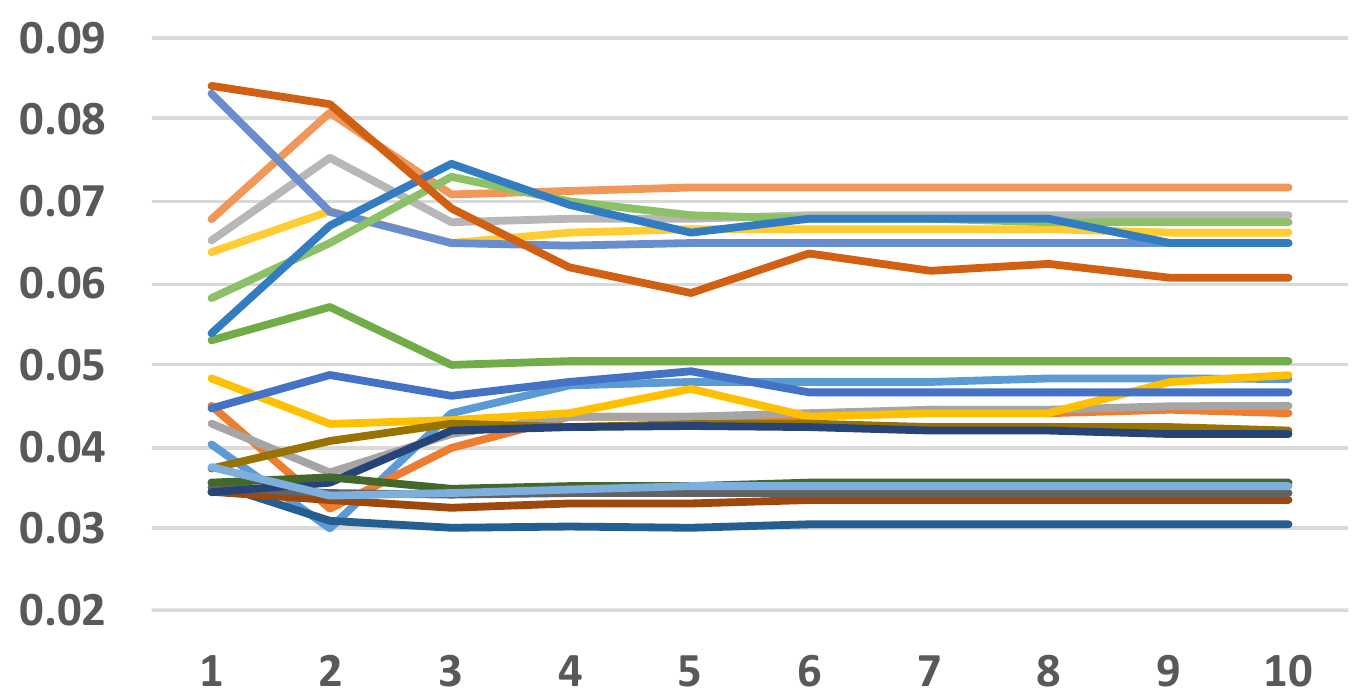}
        \caption{Job}
    \end{subfigure}    
    \begin{subfigure}[t]{1\textwidth}
        \centering
        \includegraphics[height=0.25in]{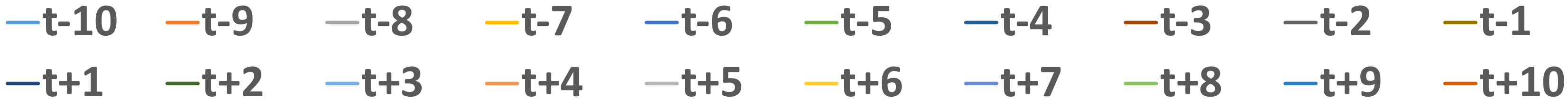}
    \end{subfigure}
    \caption{Mean attention w.r.t. epoch for our model model when the user context is included in the temporal context.}
    \label{fig:att_mean_u1}
\end{figure*}

We plot the attention mean across each position of the context tweet with respect to the epoch number. From Fig~\ref{fig:att_mean_u1}, we see that mean attention at each context position are approximately in the ballpark. Mean attention weights vary for each context position, exhibiting no relation with respect to the increase in distance (as seen in Fig~\ref{fig:att_dist}). These findings indicate the complexity of giving attention to tweets in the temporal context. Initially, we see that the mean attention weights are changing drastically indicating their sub-optimality. It is interesting to see the convergence of these weights to the optimal solution is fast (in terms of no. of epochs) in the model which uses user context when compared to the model that does not use it.

\section{Conclusions} 
\label{sec:conclusion}
We proposed a model to learn generic tweet representations which have a wide range of applications in NLP and IR field. We discovered that the principled usage of the tweets in the temporal context is an important direction in enriching the representations. We also explored learning a novel user context vector to make our model user-aware while predicting the adjacent tweets. Through experimental analysis, we identified the cases when modeling the user characteristics help enhance the embedding quality. In future, we plan to understand the application potential of the user vector learned through our approach.



\end{document}